\documentclass[runningheads]{llncs}
\usepackage[T1]{fontenc}
\usepackage{graphicx}
\graphicspath{ {./visuals/fig/} }
\usepackage{hyperref}
\usepackage{enumitem}
\usepackage{amsmath}
\usepackage[table,xcdraw]{xcolor}
\usepackage{colortbl}
\usepackage{float}
\usepackage{array}

\newcolumntype{P}[1]{>{\raggedright\arraybackslash}p{#1}}
\begin{document}
\title{Consumer-friendly EEG-based Emotion Recognition System: A Multi-scale Convolutional Neural Network Approach}
\titlerunning{Consumer-friendly EEG-based Emotion Recognition}
%
%
\author{
Tri Duc Ly\inst{1} \and
Gia H. Ngo\inst{2}
}
\authorrunning{Tri Duc \& Gia, 2024}
%
\institute{
VNU-HCM High School for the Gifted, Ho Chi Minh City, Vietnam \and
School of Electrical \& Computer Engineering, Cornell University, USA}
\maketitle 
%
%
%
\begin{abstract}
EEG is a non-invasive, safe, and low-risk method to record electrophysiological signals inside the brain. Especially with recent technology developments like dry electrodes, consumer-grade EEG devices, and rapid advances in machine learning, EEG is commonly used as a resource for automatic emotion recognition. With the aim to develop a deep learning model that can perform EEG-based emotion recognition in a real-life context, we propose a novel approach to utilize multi-scale convolutional neural networks to accomplish such tasks. By implementing feature extraction kernels with many ratio coefficients as well as a new type of kernel that learns key information from four separate areas of the brain, our model consistently outperforms the state-of-the-art TSception model in predicting valence, arousal, and dominance scores across many performance evaluation metrics.

\end{abstract}
%
%
\section{Introduction}
\label{sec:sec01}

Emotion is a fundamental part of human life. Associated with feelings, emotions are important to one’s decision-making, learning process, and many other cognitive processes \cite{_01_Damasio2001Emotion}. They are also a key tool for human interaction and communication, serving as a way for the communicator to express themself, providing information on their state of mind, feelings, motives, and intentions \cite{_02_Van2010Emerging}. Therefore, the interest in researching, learning, and further understanding human emotion and its impact has been growing, especially in the field of neuroscience \cite{_03_Ali2023Review}.  While much progress is still to be made with research regarding the matter due to the complexity behind human emotion \cite{_04_Bericat2015sociology}, many advancements have been achieved thanks to the development of technology in the field, contributing to novel approaches that researchers can use to further study human emotion. With more knowledge of emotion, a key to perceiving feelings, expression, and cognitive information of a person, there are a number of ways this can be used to improve the quality of life, one of which, is therapy. 

In recent years, a number of technological advancements have led to an increased interest in a resource that can be used for the task: electroencephalography (EEG), one of the most widely used brain imaging technologies. EEG presents a non-invasive way to measure brain electrical activities, which can then be passed through a brain-computer interface (BCI) to further process the information and identify human emotion. This surge in interest is due to the development of consumer-grade EEG devices with dry electrodes. Before this development, despite its potential, the applications of EEG-BCI systems outside of research labs are extremely sparse due to the limitations of research-grade EEG devices: (1) it is time-consuming to set up a research-grade EEG device, typically taking from 30 to 60 minutes, (2) user’s mobility is heavily restricted due to the high number of wires, and (3) the extremely high cost of the devices \cite{_05_Grummett2015Measurement}. Even though data recorded by research-grade EEG devices may provide more information and allow EEG-BCI systems to yield better results in the task of emotion recognition, consumer-grade EEG devices open up countless more research and consumer applications with their affordability, portability and simplicity, while still providing reliable results \cite{_06_Suhaimi2020EEG}.

With automatic emotion recognition using EEG signals, the use of machine learning (ML) algorithms, specifically deep learning (DL), is one of the most popular and reliable methods due to its known ability to learn non-linear patterns from complex information - an aspect that is found in EEG. Thus, we developed a DL model inspired by TSception \cite{_07_Ding2021TSception} that predicts human emotion based on EEG with reliable accuracy. Our method provides a novel approach to which a DL architecture can be used for the task.

The remainder of this paper is organized as follows. In \textbf{Section~\ref{sec:sec02}}, we give background information on emotion, EEG, EEG-based emotion recognition, and deep learning. \textbf{Section~\ref{sec:sec03}} provides the details of our materials and method by introducing the dataset and the performance evaluation metrics that we used, the experimental setup of the DL model, and the mechanism behind our model. In \textbf{Section~\ref{sec:sec04}}, we present the results of our model and analyze how these results can be interpreted. Future implications, suggestions, and a discussion regarding our study are given in \textbf{Section~\ref{sec:sec05}}. Finally, the conclusion and guidance acknowledgment are presented in \textbf{Sections~\ref{sec:sec06}} and\textbf{~\ref{sec:sec07}}.

\section{Background}
\label{sec:sec02}

\subsection{Emotion}

Emotion has long been a complicated subject of research. One of the earliest attempts to explore the mechanism behind human emotions comes from the James-Lange theory of emotion in 1884 \cite{_08_James1948What}, in which psychologist William James and psychologist Carl Lange suggested that emotion is the result of a physiological response to external stimuli (e.g., your body trembling and your heart beating rapidly would cause you to feel fear). This theory was challenged in 1927 by Walter Cannon and Phillip Bard, who argued in the Cannon-Bard theory that emotion and physiological response occur simultaneously, not consequently \cite{_09_Cannon1927James}. The Freudian theory by Sigmund Freud, one of the greatest and most well-known theories in psychology, introduced the idea that much of human behavior, including human emotion, is greatly influenced by the unconscious mind. In the 1950s, the cognitive revolution led to the development of novel theories such as the Schachter-Singer Two-Factor Theory of Emotion, which suggests that emotion consists of two components: physical arousal and a cognitive interpretation \cite{_10_Schachter1962Cognitive}. In the 1960s and the 1970s, Richard Lazarus pioneered the advancement of the cognitive appraisal theory, stating emotion is the response to an individual’s evaluation of a situation. As the 21st century approaches, advances in brain imaging like functional magnetic resonance imaging led to many studies about the brain’s role in human emotions, the emergence of the field of affective neuroscience, and the popularization of emotional intelligence. These advancements, with pioneers like Jaak Panksepp and Daniel Goleman, have led to rapid progress in extensive research on the role of emotion and the mechanism behind it. 

Emotion plays many roles in human life: (1) Emotions serve as social signals in human exchanges, conveying one’s feelings, perceptions, and situational understanding to the interaction partner \cite{_11_Pons2010Cognition}; (2) Emotion greatly influences cognitive function, which includes decision making, perception, learning, and maintaining health \cite{_01_Damasio2001Emotion}; (3) Emotion significantly affects the course of actions, execution, control, and the way one explains their actions \cite{_12_Zhu2002Emotion}. It also aids our daily lives in various ways. Most evidently, emotion is known to provide memory benefits, though it enhances memory more for negative experiences than positive ones \cite{_13_Kensinger2009Remembering}. As a social signal, it serves as a tool for humans to connect with others, navigating social interactions so that we can resonate with and care for another person \cite{_14_Grecucci2017Editorial}. The positive and negative nature of emotions can also affect human health, as well as influencing and experiencing work processes \cite{_15_Ali2017Emotion}, \cite{_16_Briner1999Neglect}. 

There is no unitary definition for emotion \cite{_17_Izard2010Many}, and the way emotion is perceived has vastly varied throughout modern science. The way people feel certain emotions can be different from how others do, and the description of an emotion one feels can vastly vary from another \cite{_18_Kuppens2010Appraisal}. P. R. Kleinginna and A. M. Kleinginna \cite{_19_Kleinginna1981categorized} famously defined emotions as “a complex set of interactions among subjective and objective factors, mediated by neural~hormonal systems, which can (a) give rise to affective experiences such as feelings of arousal, pleasure/displeasure; (b) generate cognitive processes such as emotionally relevant perceptual effects, appraisals, labeling processes; (c) activate widespread physiological adjustments to the arousing conditions; and (d) lead to behavior that is often, but not always, expressive, goal-directed, and adaptive.” Generally, emotion is described as the brain’s consistent response toward an external stimulus. To classify emotion, most approaches are either categorical or dimensional. The categorical approach classifies emotions into a group of defined emotional states, or discrete emotions. Many researchers have presented various ways to define such groups:

\begin{enumerate}[leftmargin=2cm]
    \item Kemper \cite{_20_Kemper1987How} stated that there are four primary emotions - fear, anger, depression, and satisfaction - and many secondary emotions that can be acquired via social agents like guilt, shame, pride, gratitude, love, nostalgia, and ennui.
    \item Levenson \cite{_21_Levenson2011Basic} stated that basic emotions need to meet three criteria of distinctness, hard-wiredness, and functionality, and he found six emotions that suffice: enjoyment, anger, disgust, fear, surprise, and sadness.
    \item Ekman and Cordaro \cite{_22_Ekman2011What} stated that most basic emotions share 13 common characteristics and summarized seven basic emotions: Anger, fear, surprise, sadness, disgust, contempt, and happiness.
    \item Plutchik \cite{_23_Plutchik2001Nature} proposed eight basic emotions described in a wheel model: joy, trust, fear, surprise, sadness, disgust, anger, and anticipation.
    \item Izard \cite{_24_Izard2007Basic} proposed 10 basic emotions: interest, joy, surprise, sadness, fear, shyness, guilt, anger, disgust, and contempt.
\end{enumerate}

\begin{figure}[h]
\centering
\includegraphics[width=0.5\textwidth]{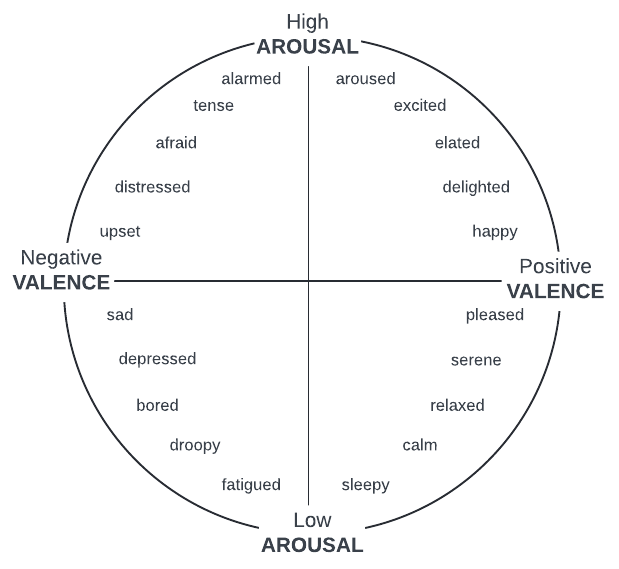}
\caption{Russell’s circumplex model of affect.}
\label{fig:fig01}
\end{figure}

\begin{figure}[h]
\centering
\includegraphics[width=0.5\textwidth]{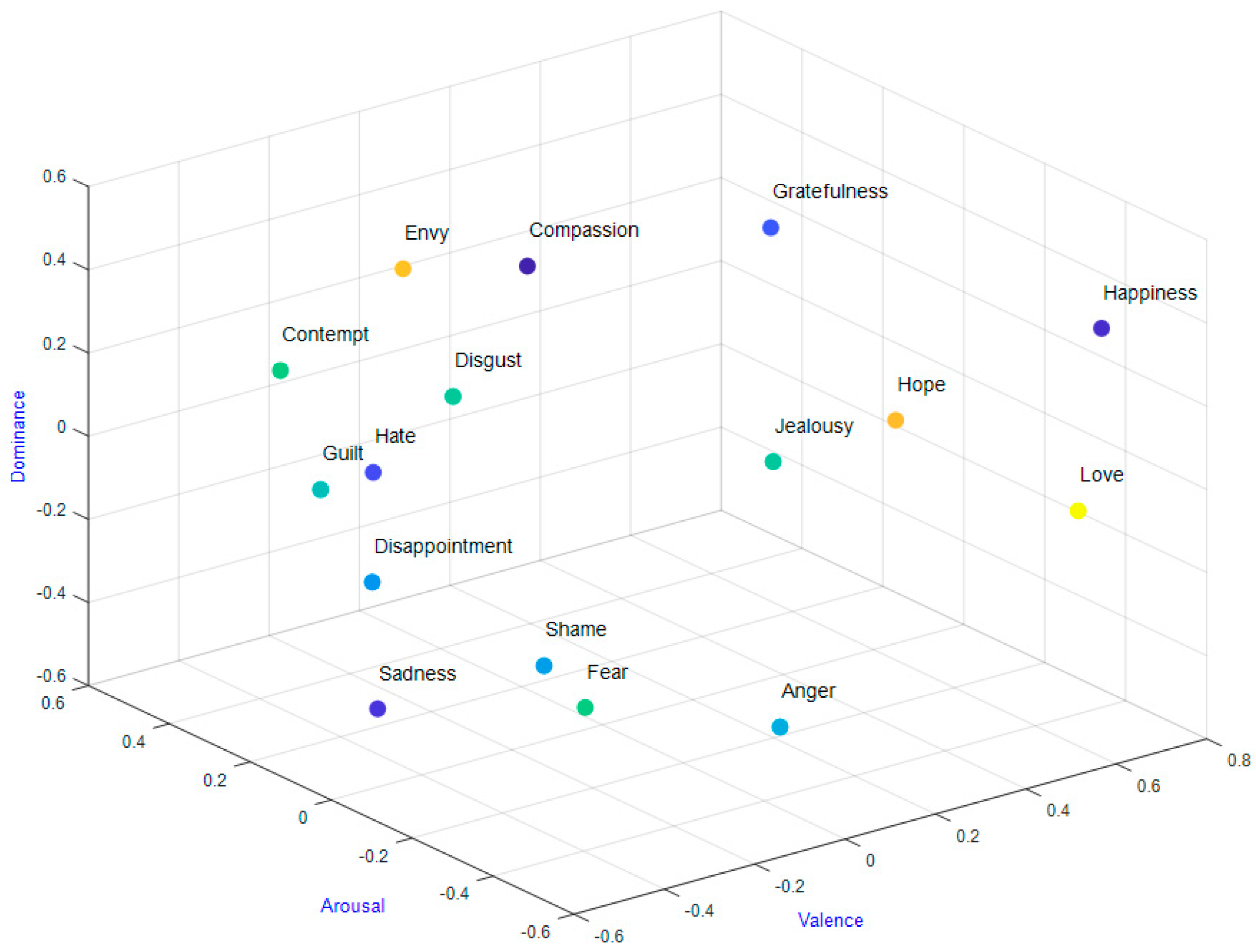}
\caption{Russell’s and Steiger’s VAD model \cite{_28_Topic2022Emotion}.}
\label{fig:fig02}
\end{figure}

On the contrary, the dimensional approach better captures the complexity of emotion by using a scale for different affective states. While this approach is still incompatible with depicting the full complexity and high-dimensional nature of emotions \cite{_25_Cowen2018Clarifying}, it allows for much easier implementation and interoperability, as well as improving accuracy for emotion classification systems. One of the most commonly used dimensional models in emotion classification systems is Russell’s circumplex model of affect, shown in \textbf{Fig.~\ref{fig:fig01}} \cite{_26_Russell1980circumplex}. It is a two-dimensional model that uses two scales - a valence scale and an arousal scale. The valence scale shows how positive or negative a person is feeling, and the arousal scale shows how high or low a person’s attention level is. Russell’s and Steiger’s extended version of this model, often regarded as the valence-arousal-dominance (VAD) model, \cite{_27_Russell1982structure} is also just as commonly used, which adds an extra dimension called dominance to show how in-controlled or submissive a person is feeling, shown in \textbf{Fig.~\ref{fig:fig02}}.

\subsection{EEG}

EEG is the electrophysiological process of measuring and collecting information on electrical fields in the brain by placing electrodes on the scalp \cite{_29_Cohen2017Where}, providing a display of electrical activities in the brain in the form of waves with varying frequencies, amplitudes, and shapes. These fields are formed when billions of electrochemical signals spontaneously pass between brain neurons in an extended space, causing the sum of these fields to be powerful enough to be recorded from outside, making EEG a completely safe and non-invasive procedure. In other words, EEG signals are the result of impulses of brain neurons. EEG does not record the activities of every individual neuron but rather captures the signals created when neurons activate at the same time.  One major limitation of EEG is that its signals, which come from cerebral activity, can be contaminated easily by other physiological signals from other electrical activities generated by the body during common occasions such as movements or eye blinks, causing EEG signals to be non-linear, non-stationary, and often overwhelmed by noise \cite{_30_Sun2021EEG}, \cite{_31_Subha2008EEG}. Raw EEG data is known to be extremely complex and challenging to interpret, requiring advanced analysis, signal processing, and feature extraction to be correctly interpreted \cite{_32_Roy2019Deep}. EEG feature is a pattern associated with a particular sensory or cognitive process \cite{_29_Cohen2017Where}, which can capture human states of mind and emotional states. Thus, EEG, especially when observed in combination with EEG features, is proven to be able to provide useful information that can reflect characteristics in response to emotional states, leading to it being a very popular choice for the task of emotion recognition \cite{_33_Musha1997Feature}, \cite{_34_Musha1997Feature}.

\begin{figure}[h]
\centering
\includegraphics[width=0.5\textwidth]{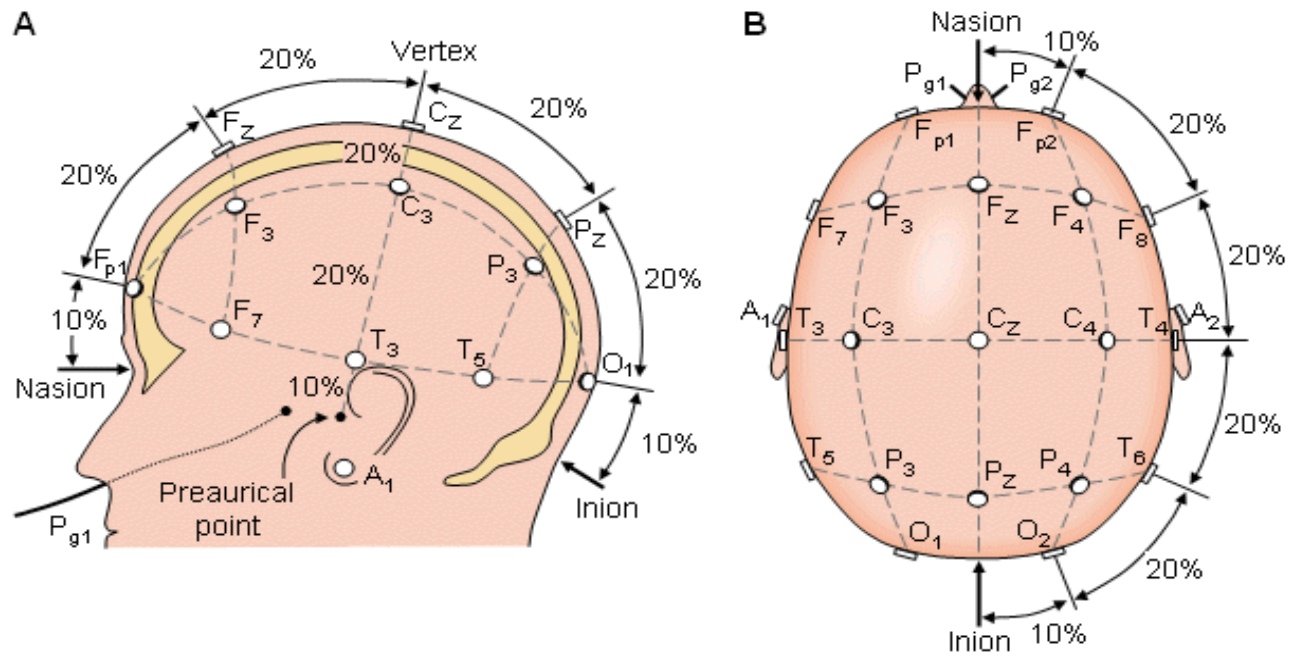}
\caption{The International 10-20 electrodes placement system (Source: \cite{_37_shriram2013eeg}).}
\label{fig:fig03}
\end{figure}

EEG electrodes record the signals of a small area surrounding them. An EEG headset can have as few as four electrodes and as many as 256 electrodes. Electrodes are often distributed all across the scalp to cover as much area as possible. Strategically placing these electrodes to yield the most satisfactory recording is one of the most important aspects of an EEG test, study, and lab research. One standardized technique to place these electrodes across the scalp is to follow the International 10-20 system, depicted in \textbf{Fig.~\ref{fig:fig03}} \cite{_35_Herwig2003Using}, \cite{_36_klem1999ten}. The International 10-20 system uses four anatomical landmarks on the scalp - the nasion (the bridge of the nose), the inion (the lowest point of the skull at the back of the head), and the left and right preauricular points (the point just in front of the ears where the upper jaw and the lower jaw meet) - as references to ensure consistent placement across EEG research:

\begin{enumerate}[leftmargin=2cm]
    \item  The nasion: The bridge of the nose. 
    \item  The inion: The lowest point of the skull at the back of the head. 
    \item  The left and right preauricular points: The two points located just in front of the ears, where the upper jaw and the lower jaw meet.
\end{enumerate}

The numbers 10 and 20 refer to the distance between one electrode to another. Every electrode that is next to an anatomical landmark will be placed 10\% of the entire distance from the front to the back or from the right to the left of the brain, and 20\% of the distance for every other electrode. A unique label is assigned to every electrode, each consisting of one or two letters and one number, representing the location of the brain where the electrode is placed. The assigned number is even if the electrode is placed in the right hemisphere of the brain, and is odd if the electrode is placed in the left hemisphere. The assigned letter (or letters) is an abbreviation of different parts of the brain: Fp for pre-frontal, F for frontal, T for temporal, O for occipital, P for parietal, and C for central.

An extension of the International 10-20 system, the 10-10 system, is also a very commonly used electrode placement system in EEG research. Instead of 10\% and 20\% intervals, the 10-10 system only uses 10\% intervals, which allows more possible positions for electrode placements, which can be more useful for purpose-specific signal recording in research.

\begin{table}[h]
\centering
\begin{tabular}{|P{3.5cm}|P{4cm}|P{4cm}|}
\hline
\textbf{Feature} & \textbf{10-20 system} & \textbf{10-10 system} \\
\hline
Number of electrodes & 19 electrodes & Typically 64 electrodes \\
\hline
Resolution & Lower spatial resolution & Higher spatial resolution \\
\hline
Electrode placement & 10\% and 20\% intervals & 10\% intervals \\
\hline
Common usage & Clinical EEG, routine monitoring & Research, studies \\
\hline
Convenience & Easier to set up and interpret & More complex, require more setup time \\
\hline
Applications & Routine diagnostics, sleep studies & Detailed brain mapping, research studies \\
\hline
\end{tabular}
\vspace{0.5em}
\caption{A comparison of the 10-20 system and the 10-10 system.}
\end{table}

\begin{figure}[h]
\centering
\includegraphics[width=0.5\textwidth]{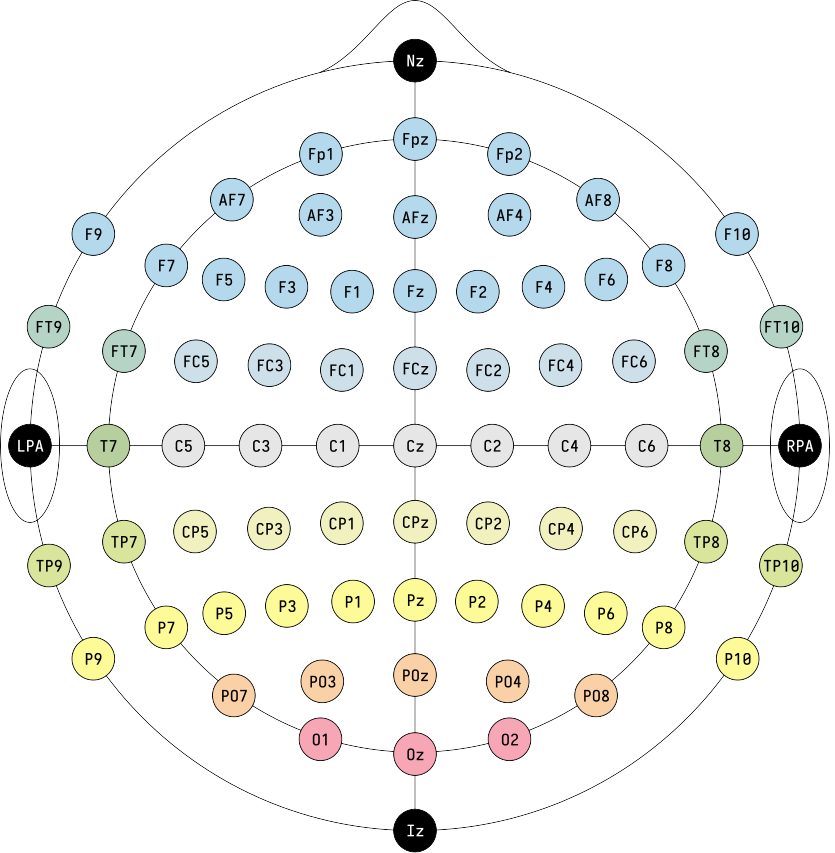}
\caption{The 10-10 electrode placement system.}
\label{fig:fig04}
\end{figure}

Thanks to the rapid development of dry electrodes and consumer-grade EEG devices from companies such as Emotiv, OpenBCI, and NeuroSky, there has been a massive boost in EEG and EEG-based emotion recognition research. Compared to the most commonly used wet electrodes, dry electrodes yield performance on a similar level to wet electrodes (often considered the gold standard for EEG electrodes) \cite{_38_Wang2012PDMS}, \cite{_39_Heijs2021Validation}, \cite{_40_Chen2014Soft}, \cite{_41_Lopez2014Dry} while still presenting a number of advantages:

\begin{enumerate}[leftmargin=2cm]
    \item Less preparation time: Dry electrodes take significantly less preparation time, as EEG headsets with wet electrodes require additional steps such as applying conductive gel \cite{_39_Heijs2021Validation}, \cite{_42_Ng2022Multi}.
    \item Better user comfort: With the use of conductive gel in wet electrodes, participants often experience discomfort, including skin irritation and inconvenient scalp preparation, especially when used for an extended time duration \cite{_38_Wang2012PDMS}, \cite{_40_Chen2014Soft}, \cite{_43_Krachunov20163D}.
    \item Easy maintenance: Since dry electrodes do not require the use of conductive gel, they are much easier to clean and require much less maintenance effort.
    \item Portability: EEG headsets with dry electrodes are more wearable and portable, as well as requiring very little scalp preparation \cite{_44_Liu2019novel}.
    \item Low cost: EEG systems with dry electrodes cost significantly less compared to systems with wet electrodes \cite{_41_Lopez2014Dry}.
\end{enumerate}

\subsection{EEG-based Emotion Recognition}

Emotion recognition refers to the task of identifying and interpreting human emotions with the help of various indications such as facial expressions, body language, or physiological signals like photoplethysmography, electromyography, and electrocardiogram (ECG). In our case, EEG-based emotion recognition uses EEG, the most commonly used option among physiological signals \cite{_06_Suhaimi2020EEG}, \cite{_45_Shu2018Review}, \cite{_46_Arthanarisamy2022Subject}. 

 In an EEG-based emotion recognition process, a study typically needs to recruit participants and select a stimulus to evoke a targeted emotion. During a recording session, the participant is often asked to wear an EEG device. The participant is then exposed to the stimulus, and the voltage fluctuations in the brain are then recorded. EEG presents an image of electrical activity in the brain. The data is then passed through a preprocessing stage, where it is denoised and artifacts are filtered out. The clean data is then analyzed and goes through the feature extraction stage, one of the most essential stages in EEG-based emotion recognition. Only then, an ML classifier is used for training and learning the EEG data, which allows it to predict the emotional state of a person, producing the final output.

\subsection{Deep Learning}

For the task of emotion recognition, artificial intelligence (AI)-based classifiers using ML algorithms have been widely used, especially in recent years, due to the rapid development of AI in general and ML specifically. While the terms AI, ML, and DL are often incorrectly used in interchange with each other, they are completely separate terms that cover different ranges of algorithms and techniques. AI includes all algorithms and techniques that allow computers to mimic human behavior \cite{_47_Janiesch2021Machine}. A lot of AI systems behave based on hard-coded statements, which limits their adaptability and ability to complete tasks with a high level of complexity. ML overcomes this by learning and improving through experience. In more technical terms, ML systems train through many iterations, compare their predictions to the actual answers to calculate a loss value, and then improve and adapt by modifying their algorithm to minimize the loss. When the loss or the error rate becomes a constant after many iterations, indicating that it is minimized, and the model’s performance stops improving, this state of an ML model is often referred to as convergence. Literature tackling emotion classification using ML often approaches it either as a classification or a regression problem. A classification task has outputs (also often regarded as labels in ML) that can be divided into a finite number of groups. An example of this would be predicting whether an email is spam or not, in which the labels can be divided into two groups: spam or non-spam. A regression task has continuous outputs that are specific values that cannot be divided into a finite number of groups. An example of this would be predicting house prices, in which the output can be any positive decimal value. There are four types of ML algorithms:

\begin{enumerate}[leftmargin=2cm]
    \item Supervised learning: These algorithms train using datasets that have pairs of input and label, and then later predict an outcome from a completely new input. Most ML algorithms fall in this group.
    \item Unsupervised learning: These algorithms train using datasets that only have outcomes and no inputs. The algorithm would then use the available data to complete a task such as association (e.g, detecting a pattern in a user’s activities in a music app and then using that pattern to suggest a song that he or she would enjoy).
    \item Semi-supervised learning: In most supervised learning tasks, it is rare to find a complete dataset. While not as commonly used, some regard to tasks where a fraction of the dataset is missing labels as semi-supervised learning.
    \item Reinforcement learning: These algorithms help a system learn to make decisions to optimize its performance using the trial-and-error learning process, which is mostly used in game theory. One well-known example of this is Google’s AlphaGo, an AI that famously defeated Go world champion Lee Sedol in 2016.
\end{enumerate}

DL builds on an ML technique called artificial neural network (ANN), which mimics how a neural network in the human brain works, amplifying or reducing signals transmitted between neurons by increasing or decreasing a weight value that is assigned to each neuron. DL systems often show better learning capabilities and accuracies compared to ML systems in many applications with high-dimensional data like text, image, video, and audio data, which is why there has been a massive interest in DL applications.

\subsubsection{Artificial Neural Network}

In the human brain, there are approximately 86 billion neurons forming 100 trillion connections to each other. To process the unending flow of information from the body, these neurons create electrical impulses non-stop to move and transmit information in a neural network. This is the basis and the inspiration for the artificial neural network: To complete a complex task, a neuron will pass on information to another neuron \cite{_48_Rosenblatt1958perceptron}, which allows it to thrive in tasks with large amounts of data that linear ML algorithms would not be able to handle. A key characteristic of ANN is its ability to learn and extract features from data, helping it make optimal decisions. An ANN has three types of layers:

\begin{enumerate}[leftmargin=2cm]
    \item Input layer: This is the layer where the data enters the ANN. Each neuron represents a distinct feature in the data. Every ANN only has one input layer.
    \item Hidden layer(s): This is where the ‘learning’ process happens in an ANN. The depth (the number of hidden layers) and the width (the number of neurons in each hidden layer) are hyperparameters that form the network architecture, which allows the network to extract and recognize complex patterns from the data. An ANN can have one or more hidden layers.
    \item Output layer: This is the final layer of the network. The output of this layer is the output of the whole network. Every ANN only has one output layer.
\end{enumerate}

\begin{figure}[h]
\centering
\includegraphics[width=0.5\textwidth]{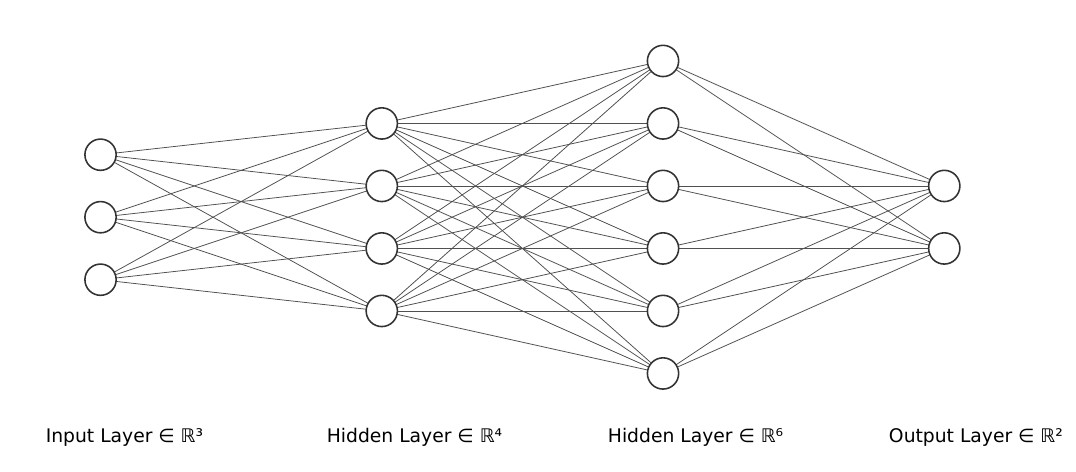}
\caption{An example of an artificial neural network, produced using NN-SVG \cite{_49_LeNail2019NN}.}
\label{fig:fig05}
\end{figure}

In an ANN, connections between neurons allow them to process information from each other with the use of weights. Each connection has its own weight, and the closer the value of the weight is to 0, the weaker the connection is. When an ANN learns via training iterations, the ANN adjusts these weights accordingly in order to minimize the error. However, for the network to process and learn complex patterns like those from real-life applications, this linear learning process would often not be good enough. To add a non-linear aspect to neurons’ connections, an ANN uses activation functions, which process the output of a neuron to determine whether it should be activated or not. Activation functions are only applied in hidden layers, and the activation function used in each hidden layer varies, making it another hyperparameter. In a linear ANN model, the activation function would simply be $f(x)=x$. For more complex ANN models, other activation functions are used, like Rectified Linear Unit (ReLU), Sigmoid, and Tanh. Finally, a bias might be applied to adjust the output independently of the input data, allowing the model to better fit the data when training. 

\subsubsection{Deep Learning}

ANN is the backbone of DL. The field of DL focuses on utilizing neural networks (NN) in a more complicated way to learn complex patterns in large datasets. DL has continuously outperformed traditional ML techniques and achieved state-of-the-art results in emotion recognition \cite{_50_Tzirakis2017End}, \cite{_51_Sharma2020Automated}, \cite{_52_Thuseethan2022Deep} thanks to its effectiveness in processing auditory and visual data. Unlike traditional ML techniques, DL removes the need for manual feature extraction and feature engineering as it can automatically learn from raw data after many iterations (often regarded in ML as epochs). DL models excel when there is a large amount of data, as their performance generally improves as more data is fed into them, making them suitable for applications with large datasets. However, a major limitation of DL models is that training them can be computationally expensive, requiring powerful GPUs, large amounts of memory, and long training time. 

\subsubsection{Convolutional Neural Network}

\begin{figure}[h]
\centering
\includegraphics[width=0.5\textwidth]{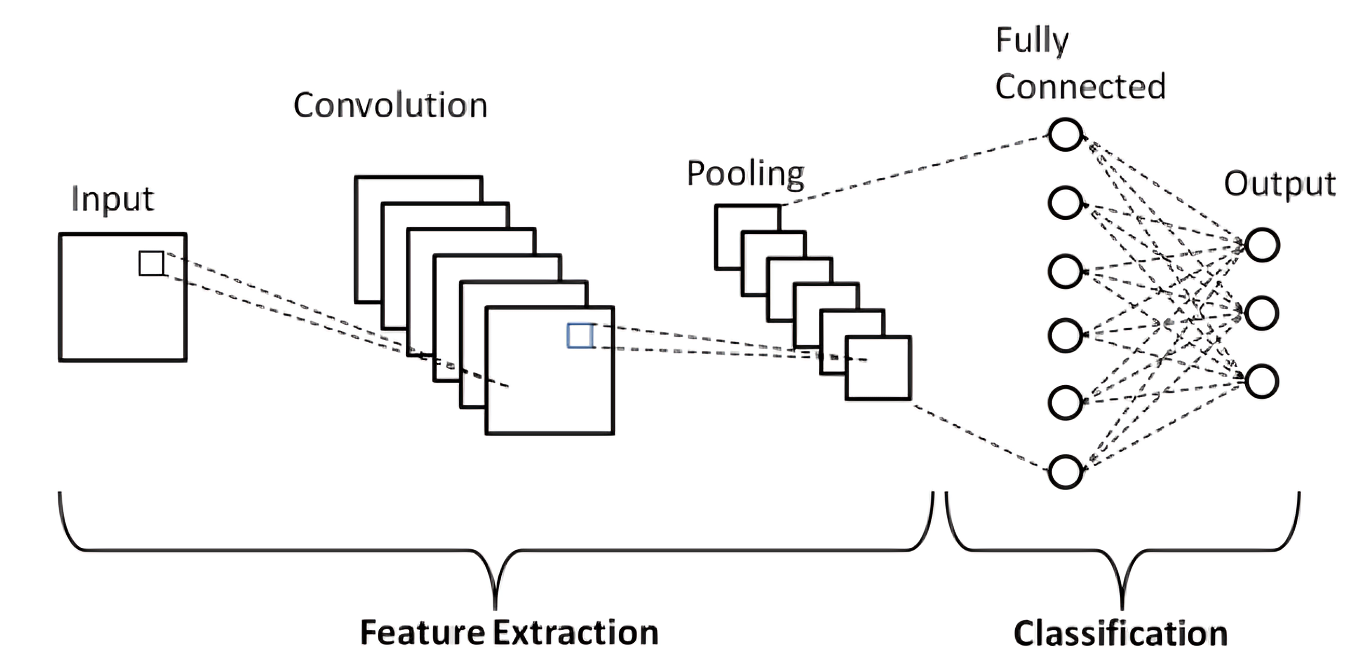}
\caption{An example of CNN \cite{_53_Phung2019High}.}
\label{fig:fig06}
\end{figure}

Convolutional neural network (CNN) is a deep learning technique that specializes in processing structured grid-like data like images. The most fundamental component of CNN is the convolutional layer. A convolutional layer has parameters called kernels or sliding windows, which act as filters sliding over input data and performing convolution operations to detect complex patterns in the data. The output of a convolutional layer is called a feature map. To make CNN models more robust to minor changes and noise, like a slight shift or a slight rotation in an image, pooling (subsampling) layers are used to reduce the width and height of the feature maps, filtering out insignificant details and retaining the most useful information. This process is often regarded in ML as downsampling. After the data is processed through many convolutional and pooling layers, the output is passed to one or several fully connected layers similar to an ANN. The fully connected layers used the features extracted to produce the final output, which in the example shown in \textbf{Fig.~\ref{fig:fig06}} would be class labels.

\section{Materials and Methods}
\label{sec:sec03}

\subsection{Dataset}

\begin{figure}[h]
\centering
\includegraphics[width=0.5\textwidth]{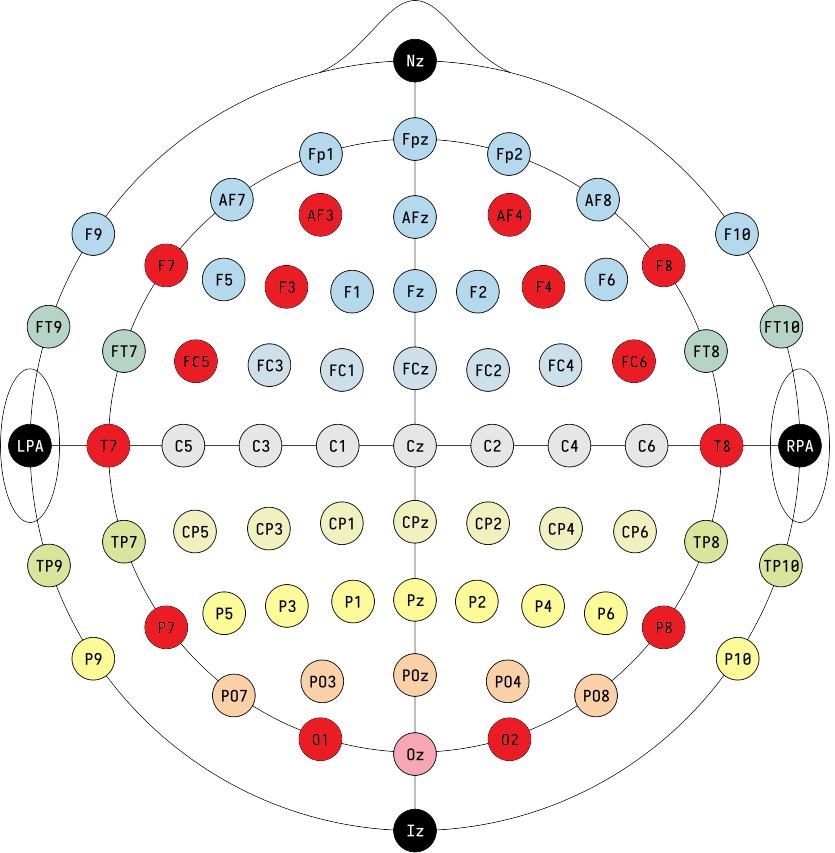}
\caption{All sensor locations used in the DREAMER dataset are colored in red.}
\label{fig:fig07}
\end{figure}

\begin{figure}[h]
\centering
\includegraphics[width=0.5\textwidth]{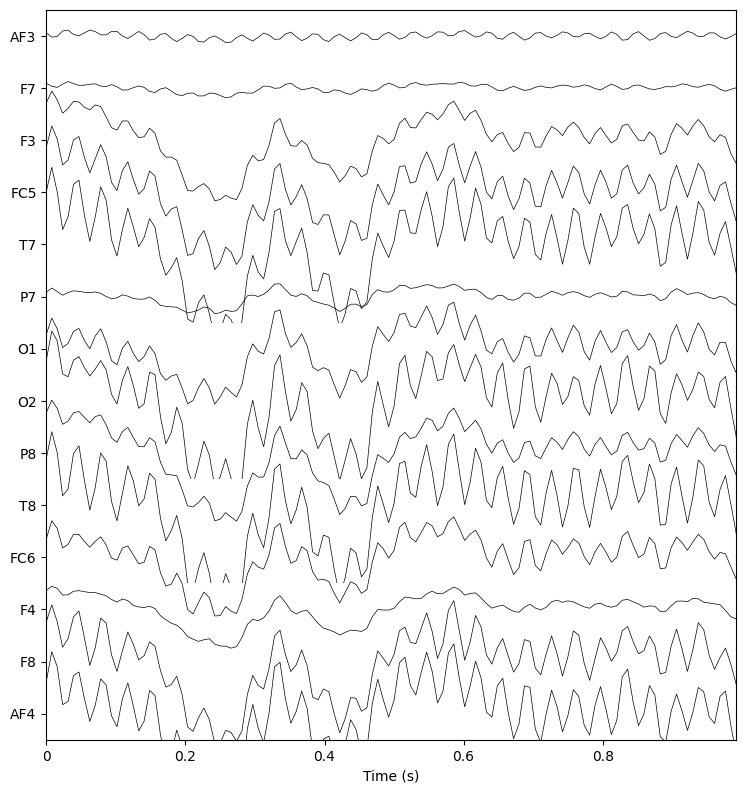}
\caption{The signal plot for one second of recording taken from the DREAMER dataset.}
\label{fig:fig08}
\end{figure}

The DREAMER dataset is a commonly used public dataset in the field of affective computing, especially for emotion recognition based on EEG and ECG. It is a multi-modal dataset that contains both EEG and ECG signals recorded during affect elicitation using audio-visual stimuli \cite{_54_Katsigiannis2018DREAMER}. The dataset includes recordings from 25 participants (14 males and 11 females) aged between 22 and 33 with an average age of 26.6 years, though technical problems caused recording from two subjects (both females) to be unsuitable for use. In emotion recognition, audio-visual stimuli (video) guarantee more valence intensity in subjects compared to visual-only stimuli (images) \cite{_55_Apicella2021EEG}. In each session, which lasted approximately one hour, each subject started by watching a neutral film clip in order to record the baseline signals and to return the emotional state of the subject to normal after watching an emotion-eliciting film clip. Every subject will watch 18 film clips with nine different targeted emotions (calmness, surprise, amusement, fear, excitement, disgust, happiness, anger, and sadness) with durations ranging from 65 to 393 seconds (the mean duration was 199 seconds). While the subjects were viewing the film clip, their EEG was being recorded at a sample rate of 128 Hz using the Emotiv EPOC wireless EEG headset, which has 16 gold-plated contact sensors placed on locations in accordance with the International 10-10 system. One mastoid sensor was placed at M1 as a ground reference point to measure the voltage for other sensors, and another mastoid sensor was placed at M2 as a feed-forward reference for reducing external electrical interference. The other 14 sensors were placed in the following locations: AF3, F7, F3, FC5, T7, P7, O1, O2, P8, T8, FC6, F4, F8, and AF4, as shown in \textbf{Fig~\ref{fig:fig07}}. After watching each film clip, the subject will be asked to evaluate their emotion (based on what they actually felt, not what they thought the targeted emotion of the film clip is) by reporting the felt valence, arousal, and dominance scores on a five-point scale for each of them. An example of how the EEG signal plot would look is shown in \textbf{Fig.~\ref{fig:fig08}}, which is the signal plot for one second of recording taken randomly from the DREAMER dataset. The recorded EEG and ECG data, the participants’ data, and their valence, arousal, and dominance evaluation were then stored in a data structure. A summary of the experiment is shown in \textbf{Table~\ref{tab:table02}}.

\begin{table}[h]
\centering
\begin{tabular}{|P{5cm}|P{7cm}|}
\hline
\multicolumn{2}{|c|}{\textbf{Audio-visual stimuli}} \\
\hline
\textbf{Number of videos} & 18 \\
\hline
\textbf{Video content} & Audio-video \\
\hline
\textbf{Video duration} & 65–393s (M=199s) \\
\hline
\multicolumn{2}{|c|}{\textbf{Experiment information}} \\
\hline
\textbf{Number of participants} & 25 (only the data of 23 participants was used) \\
\hline
\textbf{Number of males} & 14 \\
\hline
\textbf{Number of females} & 11 (only the data of 9 females was used) \\
\hline
\textbf{Age of participants} & 22–23 (M=26.6, SD=2.7) \\
\hline
\textbf{Rating scales} & Arousal, Valence, Dominance \\
\hline
\textbf{Rating values} & 1–5 \\
\hline
\textbf{Recorded signals} & 14-channel 128Hz EEG \\
\hline
\end{tabular}

\vspace{0.5em}
\caption{The experiment summary of the DREAMER dataset \cite{_54_Katsigiannis2018DREAMER}.}
\label{tab:table02}
\end{table}

Since we are developing an EEG-based emotion recognition system, only the EEG data of the participants is considered. A threshold of three was used to convert the arousal, valence, and dominance scores from 1-5 to zero and one, which means all scores with values greater than or equal to three would be converted to one, and all other values would be converted to 0. This is done to minimize the effect of the possible variation for each participant in the evaluation of elicited emotion. 

\subsection{Experimental Setup}

\subsubsection{Metrics}
\label{sec:metrics}

To compare our model to the TSCeption model, we evaluate the performance of each model using many different metrics: precision, recall, F1 score, accuracy, Matthew’s correlation coefficient (MCC), Cohen’s kappa, and area under the receiver operating characteristic curve (AUROC). All of those metrics can be calculated using the true positive (TP), true negative (TN), false positive (FP), and false negative (FN) outcomes of the models, which represent when:

\begin{enumerate}[leftmargin=2cm]
    \item TP: Both the prediction and the actual value are positive.
    \item TN: Both the prediction and the actual value are negative.
    \item FP: The predicted value is positive, but the actual value is negative.
    \item FN: The predicted value is negative, but the actual value is positive.
\end{enumerate}

Despite the most applied metrics for evaluating binary classification models being F1 score and accuracy, we evaluated the models using this wide variety of metrics, as each comes with its own strengths and limitations. Thus, for each different potential application of the model, one metric may provide a better evaluation compared to others. Precision, recall, F1 score, accuracy, and MCC can be calculated using these equations:
\begin{align}
\text{Precision} &= \frac{TP}{TP + FP} \\
\text{Recall} &= \frac{TP}{TP + FN} \\
\text{F1 score} &= \frac{2 \cdot \text{Precision} \cdot \text{Recall}}{\text{Precision} + \text{Recall}} \\
\text{Accuracy} &= \frac{TP + TN}{TP + TN + FP + FN} \\
\text{MCC} &= \frac{TP \cdot TN - FP \cdot FN}{\sqrt{(TP + FP)(TP + FN)(TN + FP)(TN + FN)}} \\
\text{AUROC} &= \frac{1}{2} \left( \frac{TP}{TP + FN} + \frac{TN}{TN + FP} \right) \\
\kappa &= \frac{2(TP \cdot TN - FP \cdot FN)}{(TP + FP)(FP + TN) + (TP + FN)(FN + TN)}
\end{align}

Precision is defined as the number of correct positive predictions divided by the total number of predicted positives. In other words, it evaluates the accuracy of a model’s positive prediction. Recall is defined as the number of correct positive predictions divided by the number of actual positive values. In our case, this is the number of ones in predicted values divided by the number of ones in labels. Combining the pair of precision and recall provides a useful evaluation of classification models, which the F1 score achieves by calculating the harmonic mean of precision and recall. Thus, the F1 score remains one of the most popular choices for evaluating classification models. However, it is worth noting that the F1 score also has a number of disadvantages that prevent it from fully reflecting the performance of an ML model: 

\begin{enumerate}[leftmargin=2cm]
    \item The F1 score can be easily affected when dealing with an imbalanced class, as it can be dominated by the majority class. 
    \item As can be seen in equation (3), the F1 score does not take into account any true negative values, which can result in misleading conclusions, especially when evaluating models with a task that might be important to understand true negatives, like certain medical tests or anomaly detections.
    \item Precision and recall have equal weights in the calculation of the F1 score, assuming equal importance between them, which results in domain knowledge not being taken into account. In many cases, this can lead to significant, misleading conclusions from the evaluation (e.g., precision might be more important in fraud detection, whereas recall might be more important in medical diagnostics). 
    \item The F1 score is sensitive to minor changes in small datasets, which can result in instability in the metric.
    \item As the F1 score is the harmonic mean of precision and recall, it is less interpretable compared to other metrics like precision, recall, or accuracy.
\end{enumerate}

Accuracy, like the F1 score, is one of the most popular choices for evaluating an ML model. It is highly utilized, especially among models built for non-complex tasks, due to its simplicity in calculating and interpretability. Accuracy represents the number of correct predictions divided by the total number of predictions. This simplicity, on the contrary, also brings many limitations to the metric. Accuracy is only effective when the dataset has a balanced class, as in a dataset with class imbalance, a model can achieve a good accuracy score simply by having all predictions as the majority class, resulting in ineffective training. Moreover, since accuracy only considers true positives and negatives, the metric lacks information that would otherwise be reflected in other metrics like precision and recall, which might lead to incomprehensive conclusions.

MCC, ranging from -1 to +1, represents the correlation coefficient between the observed and predicted binary classifications. Since MCC takes into account all four categories of the confusion matrix (TP, FP, TN, and FN), MCC is considered a balanced metric, particularly useful for imbalanced datasets [56]. An MCC of +1 indicates perfect predictions, 0 indicates a prediction capability equal to completely random, and -1 indicates complete disagreement between predictions and actual outcomes.

AUROC (sometimes simply referred to as AUC) measures the ability of a binary classifier to distinguish between positive and negative classes across different thresholds. It is calculated by finding the area under the ROC curve, which plots the true positive rate (TPR) against the false positive rate (FPR). AUROC ranges from 0 to 1, with 1 indicating a perfect classifier, 0.5 indicating a completely random classification, and any value <0.5 indicating a worse capability than random guessing.

Cohen’s kappa measures the agreement between two raters (in our case, the DL model and the labels) while taking into account the possibility of agreement occurring by chance. This metric is widely used to assess inter-rater reliability and in classification tasks with more than two classes. The kappa ranges from -1 to 1, with +1 indicating perfect agreement beyond chance, 0 indicating that agreement is no better than chance, and any value <0 indicating systematic disagreement.

\subsubsection{Software, Programming Language, and Libraries}

All data pre-processing, analysis, and evaluation were conducted and implemented using Python 3.10.12 and the TorchEEG library on Google Colab. Developed by Zhang et al. \cite{_57_Zhang2024TorchEEGEMO}, TorchEEG is the first and one of the most widely used DL toolboxes for EEG-based emotion recognition, which was developed to aid researchers with EEG and ML research. TorchEEG separates the DL workflow into five modules: \texttt{datasets}, \texttt{transforms}, \texttt{model\_selection}, \texttt{models}, and \texttt{trainers}, each with its own plug-and-play functionalities for EEG-based emotion recognition. These modules incorporate state-of-the-art algorithms in the field, one of which we utilized for result comparison was the TSception model, as well as provide unique adaptations of DL models like transformers and diffusion models. Since the experimental protocol of DREAMER was implemented using the MATLAB environment \cite{_54_Katsigiannis2018DREAMER}, the SciPy library is used for loading the DREAMER dataset into the Google Colab notebook. We used the Scikit-learn library to implement some ML algorithms, which include logistic regression, random forest, and support vector machine, to do the same task with the DREAMER dataset for comparison. Our multi-scale CNN was implemented using PyTorch Lightning, a popular wrapper for PyTorch developed by Meta AI, and was heavily inspired by TSception \cite{_07_Ding2021TSception}. Compared to other commonly used libraries like Google’s TensorFlow or PyTorch, PyTorch Lightning has three strengths: (1) it minimizes boilerplate code for training loops, (2) it has a great advantage in distributed GPU training for scalability, and (3) it offers built-in features for model checkpoints, logging, and experiment tracking. These strengths allow developers and researchers to focus on the experimental aspects of a DL model, which is why we chose to utilize PyTorch Lightning for the development of this multi-scale CNN model.

\subsection{Methodology}

This section presents the mechanism behind our approach. In our EEG-based emotion recognition workflow, we first preprocess the raw EEG data so to increase the model performance, then the key features of the data are extracted using a multi-scale CNN, which is then finally passed into a classifier to perform emotion classification.

\subsubsection{Data Preprocess}

\begin{figure}[h]
\centering
\includegraphics[width=0.5\textwidth]{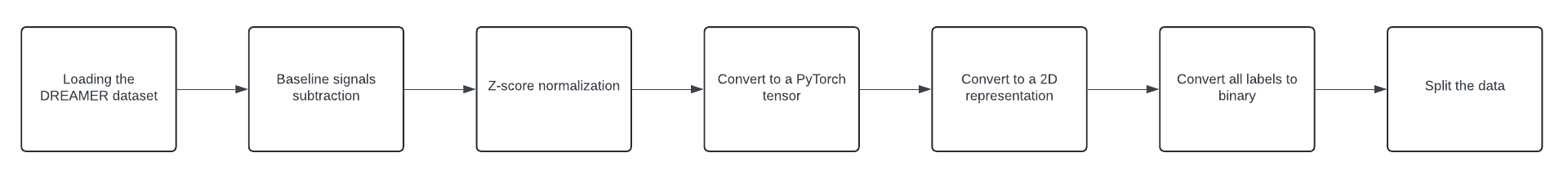}
\caption{Data Processing Overview.}
\label{fig:fig09}
\end{figure}

In EEG-based emotion recognition, preprocessing raw EEG data is one of the most essential steps for several reasons. Raw EEG signals often contain a lot of noise from body movement, electrical interference, and external environmental factors, and can be overwhelmed very easily by other artifacts \cite{_30_Sun2021EEG}. EEG signals can also vary across individuals due to the many differences, like head size. Thus, it is crucial to preprocess the raw EEG data from the DREAMER dataset. This would not only prevent the EEG data from being too noisy and complex but also allow the DL model to improve its reliability and computational expense and generalize well across different subjects, making it more suitable for real-world EEG-based emotion recognition applications \cite{_58_li2021testing}. A high-level overview of the data preprocessing step is shown in \textbf{Fig~\ref{fig:fig09}}. First, we subtracted the baseline signals from the emotional signals, then normalized the data using Z-score normalization; next, we converted the data from a MAT file to PyTorch tensor to a two-dimensional representation, converting all labels to a binary form of zero or one, and finally split the data in two ways: five-fold cross-validation and train-validation-test.

After loading the DREAMER dataset into the Google Colab notebook environment using the SciPy library, all of the data preprocessing was done using built-in functions from TorchEEG’s transforms module. First, the \texttt{BaselineRemoval} method was used to subtract the baseline signals from the experimental signals (or emotional signals). The baseline signals represent the electrical activity of the brain in a neutral or resting state with no elicited emotional state. In the DREAMER dataset, these signals are achieved by showing a neutral video clip used in the work of Gabert-Quillen et al. \cite{_59_Gabert2014Ratings}. By subtracting the baseline signals from the emotional signals, the noisy signals from background activities and signals that are unrelated to the emotional stimulus are removed, highlighting the brain’s neural activity in response to the emotion-eliciting film clips. Moreover, every individual has a distinct neutral brain activity due to various factors, such as mood, cognitive state, and anatomical differences \cite{_60_Boly2007Baseline}, \cite{_61_Fathima2023Hierarchical}. Subtracting the baseline signals will allow our EEG-based emotion recognition system to better generalize across different sessions, individual users, and stimuli in real-life applications.

Next, the \texttt{MeanStdNormalize} method was utilized to perform Z-score normalization on the EEG data. This normalization technique scales every value in a dataset so that the mean of all values is zero and the standard deviation is one. The formula for Z-score normalization is presented in the Equation:

\begin{equation}
    Z=\frac{x-\mu}{\sigma}
\end{equation}

where $Z$ is the calculated value, $x$ is the original value, $\mu$ is the mean of the data, and $\sigma$ is the standard deviation of the data. EEG typically consists of multiple channels, one for each electrode, that can have different baseline levels and variances. Z-score normalization ensures that signals from different channels and individuals are comparable. This allows the EEG-based emotion recognition system to be more robust, as well as helping ML and DL models to perform better and converge faster since the input features now have similar scales.

Furthermore, we also used the \texttt{ToTensor} and the \texttt{To2d} methods to convert the EEG data to a PyTorch tensor, then to a two-dimensional EEG signal representation, with the electrode index as the row index, and the temporal index as the column index (an additional dimension is also appended only because PyTorch requires an extra dimension to perform convolution on a two-dimensional tensor). This is a mandatory step for our model, as we utilized CNN for emotion recognition, which is designed to capture spatial patterns and features, thus the need to treat the EEG data as a two-dimensional representation. Moreover, by arranging the electrode index as the row index, the spatial relationship between electrode placements is preserved. This is beneficial to our model as the spatial configuration of the electrodes \cite{_62_Wang2022Spatial}. With the temporal index as the column index, the temporal information is also preserved, which can be crucial for real-life applications, as EEG-based emotion recognition systems need to recognize emotional states that are unfolding over time.

After that, we utilized the \texttt{Binary} method to convert all labels (the valence, arousal, and dominance scores) from a scale of 1-5 to a value of zero or one. As mentioned in \textbf{Section~\ref{sec:metrics}}, this conversion minimizes the effect of the possible variation for each participant in the evaluation of elicited emotion, helping the model generalize in real-life applications.

Finally, we split the data in two ways: five-fold cross-validation and train-validation-test split. For five-fold cross-validation, we split the partitioned dataset into five nearly equally-sized subsets (or folds) at the dimension of trials, with four subsets being the training data and one being the test set. For the train-validation-test split, we applied the standard 80-20 split ratio twice, which means the ratio between the train, validation, and test sets is 64:16:20. The model’s performance on each split method is evaluated separately. This is done to ensure that the performance evaluation of the model is as objective and comprehensive as possible.

\subsubsection{Multi-scale Convolutional Neural Network Model}

\begin{figure}[h]
\centering
\includegraphics[width=0.5\textwidth]{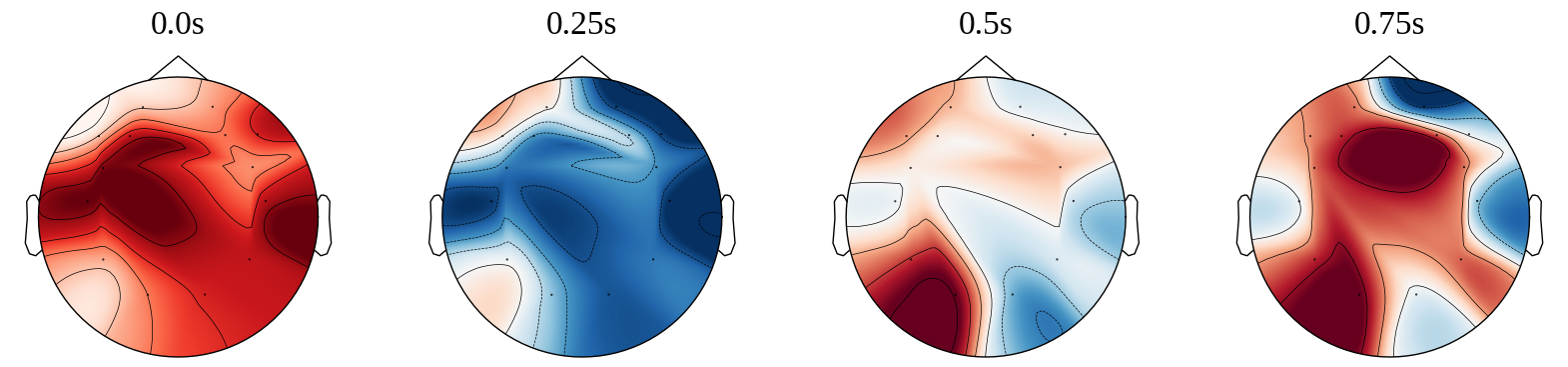}
\caption{A topographic map of the raw EEG signal across 0.75s.}
\label{fig:fig10}
\end{figure}

The architecture of our multi-scale CNN model is inspired by TSception, developed by Ding et al. \cite{_07_Ding2021TSception}. We utilized multi-scale 1D convolutional temporal kernels (T kernels) in order to extract time-frequency representations and features in EEG data. This is important due to how rapid brain activities can vary as a person is being exposed to the stimulus, which can be seen from an example shown in \textbf{Fig~\ref{fig:fig10}}. However, while TSception used the ratio coefficients of \texttt{[0.5, 0.25, 0.125]} from their hypothesis that multi-scale T kernels can better learn dynamic frequency representations in EEG, we decided to use the ratio coefficients of \texttt{[0.5, 0.25, 0.125, 0.0625, 0.03125]} so that the model can learn even more diverse representations. Although the higher-level T kernels with smaller ratio coefficients reduce the lengths of the convolutional kernels, allowing the model to learn more diverse representations, one limitation of this approach is that it can lead to the model being more computationally expensive and taking more time to run. Kernels with high ratio coefficients will learn low-frequency and long-term representations, and vice versa. In PyTorch Lightning, while there is no built-in method of implementing a one-dimensional convolutional kernel, we implemented this by setting the stride of the kernel to one. Average pooling is applied after every convolutional operation to prevent noise in the EEG data from affecting the performance of the model. All outputs of every level of T kernels would then be concatenated along the feature dimension. 

In the asymmetric spatial layer, TSception implemented two types of spatial kernels: a global kernel and a hemisphere kernel. Similar to their implementation, the global kernel has a size of ($c$, 1), where $c$ is the number of channels, and in our case, with the DREAMER dataset, 14 channels. This is done so that the length of the kernel equals the channel dimension of the EEG data, allowing the kernel to learn global spatial information through the whole scalp. The hemisphere kernel has the function of extracting the features and the relations between the left and right hemispheres of the brain. The size and step of this kernel are both ($0.5\cdot c$, 1), which allows the kernel to learn the patterns of the two hemispheres without overlapping, essentially extracting key features from them separately. However, we expanded on this approach by implementing an additional type of spatial kernel, which, instead of learning the patterns from two sides of the brain separately, the kernel extracts information from four separate areas. Essentially, we used the same implementation approach to TSception, where we set both the size and step of this kernel ($0.25\cdot c$, 1), allowing the model to learn key neural features of the brain in four parts. For both the hemisphere and our novel kernel to be well applied, the sequence of channels in the input EEG samples needs to be arranged in a specific way so that the step of the kernel would extract information from the intended areas of the brain. In order to achieve this, we arranged the channels in the input EEG data in an anti-clockwise order.

Before the output of the multi-scale CNN is finally passed onto the fully connected layer to classify the emotional states, the learned information of the three types of spatial kernels is then fused in a high-level fusion layer. Here, we implemented the proposed approach of TSception to use a 1D convolutional layer with a kernel size of (8, 1) (changed from the choice of (3, 1) from TSception due to our implementation of an additional type of spatial kernel) to fuse the learned information along the spatial dimension.

\subsubsection{Classifier}

After the key information from the EEG data is extracted using the multi-scale CNN, it is then passed into a classifier to perform emotion classification, producing the final output. This classifier essentially works like the hidden and output layers of an ANN. For each score type, the number of classes in the output would be two, making it a binary classifier.


\section{Results}
\label{sec:sec04}

In this section, we report and compare the results of our multi-scale CNN model against the state-of-the-art TSception model in terms of precision, recall, F1 score, accuracy, MCC, AUROC, and Cohen’s kappa, as well as the results of both models in five-fold cross-validation. Across these metrics, our proposed method consistently outperforms TSception, demonstrating a significant improvement in EEG-based emotion recognition with the DREAMER dataset. The results of each model across different metrics are presented in \textbf{Table 3-8}.

\begin{table}[h]
\centering
\label{tab:valence_results}
\resizebox{\textwidth}{!}{%
\begin{tabular}{|P{3.5cm}|>{\centering\arraybackslash}p{1.8cm}|>{\centering\arraybackslash}p{1.8cm}|>{\centering\arraybackslash}p{1.8cm}|>{\centering\arraybackslash}p{1.8cm}|>{\centering\arraybackslash}p{1.8cm}|>{\centering\arraybackslash}p{2cm}|}
\hline
\textbf{} & \cellcolor[HTML]{F4CCCC}\textbf{Fold 0} & \cellcolor[HTML]{F4CCCC}\textbf{Fold 1} & \cellcolor[HTML]{F4CCCC}\textbf{Fold 2} & \cellcolor[HTML]{F4CCCC}\textbf{Fold 3} & \cellcolor[HTML]{F4CCCC}\textbf{Fold 4} & \cellcolor[HTML]{F4CCCC}\textbf{Average} \\
\hline
\textbf{Tsception} & \cellcolor[HTML]{F4CCCC}75.61 & \cellcolor[HTML]{F4CCCC}78.69 & \cellcolor[HTML]{F4CCCC}78.26 & \cellcolor[HTML]{F4CCCC}78.75 & \cellcolor[HTML]{F4CCCC}78.38 & \cellcolor[HTML]{F4CCCC}77.94 \\
\hline
\textbf{Our model} & \cellcolor[HTML]{F4CCCC}\textbf{76.16} & \cellcolor[HTML]{F4CCCC}\textbf{79.33} & \cellcolor[HTML]{F4CCCC}\textbf{79.82} & \cellcolor[HTML]{F4CCCC}\textbf{78.84} & \cellcolor[HTML]{F4CCCC}\textbf{78.58} & \cellcolor[HTML]{F4CCCC}\textbf{78.55} \\
\hline
\textbf{Improvement} & 0.55 & 0.64 & 1.56 & 0.09 & 0.20 & 0.61 \\
\hline
\end{tabular}%
}

\vspace{0.5em}
\caption{The testing accuracies across five folds for valence score in five-fold cross-validations.}
\end{table}

\begin{table}[h]
\centering
\label{tab:arousal_results}
\resizebox{\textwidth}{!}{%
\begin{tabular}{|P{3.5cm}|>{\centering\arraybackslash}p{1.8cm}|>{\centering\arraybackslash}p{1.8cm}|>{\centering\arraybackslash}p{1.8cm}|>{\centering\arraybackslash}p{1.8cm}|>{\centering\arraybackslash}p{1.8cm}|>{\centering\arraybackslash}p{2cm}|}
\hline
\textbf{} & \cellcolor[HTML]{FFF2CC}\textbf{Fold 0} & \cellcolor[HTML]{FFF2CC}\textbf{Fold 1} & \cellcolor[HTML]{FFF2CC}\textbf{Fold 2} & \cellcolor[HTML]{FFF2CC}\textbf{Fold 3} & \cellcolor[HTML]{FFF2CC}\textbf{Fold 4} & \cellcolor[HTML]{FFF2CC}\textbf{Average} \\
\hline
\textbf{Tsception} & \cellcolor[HTML]{FFF2CC}\textbf{83.85} & \cellcolor[HTML]{FFF2CC}86.43 & \cellcolor[HTML]{FFF2CC}\textbf{87.07} & \cellcolor[HTML]{FFF2CC}86.11 & \cellcolor[HTML]{FFF2CC}85.41 & \cellcolor[HTML]{FFF2CC}85.774 \\
\hline
\textbf{Our model} & \cellcolor[HTML]{FFF2CC}83.59 & \cellcolor[HTML]{FFF2CC}\textbf{86.81} & \cellcolor[HTML]{FFF2CC}86.39 & \cellcolor[HTML]{FFF2CC}\textbf{87.27} & \cellcolor[HTML]{FFF2CC}\textbf{87.65} & \cellcolor[HTML]{FFF2CC}\textbf{86.342} \\
\hline
\textbf{Improvement} & -0.26 & 0.38 & -0.68 & 1.16 & 2.24 & 0.568 \\
\hline
\end{tabular}%
}

\vspace{0.5em}
\caption{The testing accuracies across five folds for arousal score in five-fold cross-validations.}
\end{table}

\begin{table}[h]
\centering
\label{tab:dominance_results}
\resizebox{\textwidth}{!}{%
\begin{tabular}{|P{3.5cm}|>{\centering\arraybackslash}p{1.8cm}|>{\centering\arraybackslash}p{1.8cm}|>{\centering\arraybackslash}p{1.8cm}|>{\centering\arraybackslash}p{1.8cm}|>{\centering\arraybackslash}p{1.8cm}|>{\centering\arraybackslash}p{2cm}|}
\hline
\textbf{} & \cellcolor[HTML]{D9EAD3}\textbf{Fold 0} & \cellcolor[HTML]{D9EAD3}\textbf{Fold 1} & \cellcolor[HTML]{D9EAD3}\textbf{Fold 2} & \cellcolor[HTML]{D9EAD3}\textbf{Fold 3} & \cellcolor[HTML]{D9EAD3}\textbf{Fold 4} & \cellcolor[HTML]{D9EAD3}\textbf{Average} \\
\hline
\textbf{Tsception} & \cellcolor[HTML]{D9EAD3}85.46 & \cellcolor[HTML]{D9EAD3}87.98 & \cellcolor[HTML]{D9EAD3}88.20 & \cellcolor[HTML]{D9EAD3}\textbf{87.79} & \cellcolor[HTML]{D9EAD3}87.46 & \cellcolor[HTML]{D9EAD3}87.378 \\
\hline
\textbf{Our model} & \cellcolor[HTML]{D9EAD3}\textbf{85.92} & \cellcolor[HTML]{D9EAD3}\textbf{88.13} & \cellcolor[HTML]{D9EAD3}\textbf{88.71} & \cellcolor[HTML]{D9EAD3}88.76 & \cellcolor[HTML]{D9EAD3}\textbf{88.42} & \cellcolor[HTML]{D9EAD3}\textbf{87.988} \\
\hline
\textbf{Improvement} & 0.46 & 0.15 & 0.51 & 0.97 & 0.96 & 0.61 \\
\hline
\end{tabular}%
}

\vspace{0.5em}
\caption{The testing accuracies across five folds for dominance score in five-fold cross-validations.}
\end{table}

\begin{table}[h]
\centering
\label{tab:valence_metrics}
\resizebox{\textwidth}{!}{%
\begin{tabular}{|P{3.5cm}|>{\centering\arraybackslash}p{2cm}|>{\centering\arraybackslash}p{2cm}|>{\centering\arraybackslash}p{2cm}|>{\centering\arraybackslash}p{2cm}|>{\centering\arraybackslash}p{2cm}|>{\centering\arraybackslash}p{2cm}|>{\centering\arraybackslash}p{2cm}|}
\hline
\textbf{} & \cellcolor[HTML]{F4CCCC}\textbf{Precision} & \cellcolor[HTML]{F4CCCC}\textbf{Recall} & \cellcolor[HTML]{F4CCCC}\textbf{F1 score} & \cellcolor[HTML]{F4CCCC}\textbf{Accuracy} & \cellcolor[HTML]{F4CCCC}\textbf{MCC} & \cellcolor[HTML]{F4CCCC}\textbf{AUROC} & \cellcolor[HTML]{F4CCCC}\textbf{Kappa} \\
\hline
\textbf{Tsception} & \cellcolor[HTML]{F4CCCC}75.29\% & \cellcolor[HTML]{F4CCCC}73.20\% & \cellcolor[HTML]{F4CCCC}73.82\% & \cellcolor[HTML]{F4CCCC}75.98\% & \cellcolor[HTML]{F4CCCC}48.44\% & \cellcolor[HTML]{F4CCCC}82.84\% & \cellcolor[HTML]{F4CCCC}47.90\% \\
\hline
\textbf{Our model} & \cellcolor[HTML]{F4CCCC}\textbf{75.96\%} & \cellcolor[HTML]{F4CCCC}\textbf{74.70\%} & \cellcolor[HTML]{F4CCCC}\textbf{75.16\%} & \cellcolor[HTML]{F4CCCC}\textbf{76.87\%} & \cellcolor[HTML]{F4CCCC}\textbf{50.65\%} & \cellcolor[HTML]{F4CCCC}\textbf{83.68\%} & \cellcolor[HTML]{F4CCCC}\textbf{50.42\%} \\
\hline
\textbf{Improvement} & 0.67\% & 1.50\% & 1.34\% & 0.89\% & 2.21\% & 0.84\% & 2.52\% \\
\hline
\end{tabular}%
}

\vspace{0.5em}
\caption{The precision, recall, F1 score, accuracy, MCC, AUROC, and Cohen’s kappa of our model and the TSception model for valence scores.}
\end{table}
\begin{table}[h]
\centering
\label{tab:arousal_metrics}
\resizebox{\textwidth}{!}{%
\begin{tabular}{|P{3.5cm}|>{\centering\arraybackslash}p{2cm}|>{\centering\arraybackslash}p{2cm}|>{\centering\arraybackslash}p{2cm}|>{\centering\arraybackslash}p{2cm}|>{\centering\arraybackslash}p{2cm}|>{\centering\arraybackslash}p{2cm}|>{\centering\arraybackslash}p{2cm}|}
\hline
\textbf{} & \cellcolor[HTML]{FFF2CC}\textbf{Precision} & \cellcolor[HTML]{FFF2CC}\textbf{Recall} & \cellcolor[HTML]{FFF2CC}\textbf{F1-score} & \cellcolor[HTML]{FFF2CC}\textbf{Accuracy} & \cellcolor[HTML]{FFF2CC}\textbf{MCC} & \cellcolor[HTML]{FFF2CC}\textbf{AUROC} & \cellcolor[HTML]{FFF2CC}\textbf{Kappa} \\
\hline
\textbf{Tsception} & \cellcolor[HTML]{FFF2CC}78.29\% & \cellcolor[HTML]{FFF2CC}\textbf{75.90\%} & \cellcolor[HTML]{FFF2CC}76.96\% & \cellcolor[HTML]{FFF2CC}84.22\% & \cellcolor[HTML]{FFF2CC}54.13\% & \cellcolor[HTML]{FFF2CC}87.63\% & \cellcolor[HTML]{FFF2CC}53.97\% \\
\hline
\textbf{Our model} & \cellcolor[HTML]{FFF2CC}\textbf{81.15\%} & \cellcolor[HTML]{FFF2CC}74.72\% & \cellcolor[HTML]{FFF2CC}\textbf{77.10\%} & \cellcolor[HTML]{FFF2CC}\textbf{85.24\%} & \cellcolor[HTML]{FFF2CC}\textbf{55.50\%} & \cellcolor[HTML]{FFF2CC}\textbf{88.31\%} & \cellcolor[HTML]{FFF2CC}\textbf{54.47\%} \\
\hline
\textbf{Improvement} & 2.86\% & -1.18\% & 0.14\% & 1.02\% & 1.37\% & 0.68\% & 0.50\% \\
\hline
\end{tabular}%
}

\vspace{0.5em}
\caption{The precision, recall, F1 score, accuracy, MCC, AUROC, and Cohen’s kappa of our model and the TSception model for arousal scores.}
\end{table}

\begin{table}[h]
\centering
\label{tab:dominance_metrics}
\resizebox{\textwidth}{!}{%
\begin{tabular}{|P{3.5cm}|>{\centering\arraybackslash}p{2cm}|>{\centering\arraybackslash}p{2cm}|>{\centering\arraybackslash}p{2cm}|>{\centering\arraybackslash}p{2cm}|>{\centering\arraybackslash}p{2cm}|>{\centering\arraybackslash}p{2cm}|>{\centering\arraybackslash}p{2cm}|}
\hline
\textbf{} & \cellcolor[HTML]{D9EAD3}\textbf{Precision} & \cellcolor[HTML]{D9EAD3}\textbf{Recall} & \cellcolor[HTML]{D9EAD3}\textbf{F1-score} & \cellcolor[HTML]{D9EAD3}\textbf{Accuracy} & \cellcolor[HTML]{D9EAD3}\textbf{MCC} & \cellcolor[HTML]{D9EAD3}\textbf{AUROC} & \cellcolor[HTML]{D9EAD3}\textbf{Kappa} \\
\hline
\textbf{Tsception} & \cellcolor[HTML]{D9EAD3}79.08\% & \cellcolor[HTML]{D9EAD3}\textbf{75.03\%} & \cellcolor[HTML]{D9EAD3}\textbf{76.74\%} & \cellcolor[HTML]{D9EAD3}86.07\% & \cellcolor[HTML]{D9EAD3}53.95 & \cellcolor[HTML]{D9EAD3}88.21 & \cellcolor[HTML]{D9EAD3}\textbf{53.58} \\
\hline
\textbf{Our model} & \cellcolor[HTML]{D9EAD3}\textbf{81.44\%} & \cellcolor[HTML]{D9EAD3}73.17\% & \cellcolor[HTML]{D9EAD3}76.12\% & \cellcolor[HTML]{D9EAD3}\textbf{86.58\%} & \cellcolor[HTML]{D9EAD3}\textbf{53.98} & \cellcolor[HTML]{D9EAD3}\textbf{88.74} & \cellcolor[HTML]{D9EAD3}52.59 \\
\hline
\textbf{Improvement} & 2.36\% & -1.86\% & -0.62\% & 0.51\% & 0.03 & 0.53 & -0.99 \\
\hline
\end{tabular}%
}

\vspace{0.5em}
\caption{The precision, recall, F1 score, accuracy, MCC, AUROC, and Cohen’s kappa of our model and the TSception model for dominance scores.}
\end{table}

As shown in \textbf{Table 3-5}, when evaluating valence scores, our multi-scale CNN model outperforms TSception in all metrics; when evaluating arousal scores, our model outperforms TSception in every metric except recall; and when evaluating dominance scores, our model outperforms TSception in precision, accuracy, MCC, and AUROC. These results indicate that our model not only surpasses TSception in traditional metrics like accuracy and F1 score but also presents significant improvements in more nuanced metrics like MCC and AUROC. Moreover, as shown in \textbf{Table 6-8}, even when comparing test accuracies from five-fold cross-validation, our method achieves substantial improvements from the results of the TSception model. When the average test accuracy of the five folds is calculated, our model shows better performance in all three valence, arousal, and dominance scores. Interestingly, for predicting valence scores, both these evaluation methods show that our model outperforms TSception in all metrics and in all five folds, indicating that our model is objectively better than TSception at recognizing valence in EEG-based emotion recognition. For the arousal score, while TSception presents a better result in recall, our model still obtained a higher F1 score, indicating a better balance between precision and recall, as well as a greater proficiency in identifying positive instances (when a person is feeling positive/pleased). Finally, for the dominance score, our model still outperforms TSception in most metrics, including precision, accuracy, MCC, and AUROC.

Across all three valence, arousal, and dominance scores, our model consistently outperforms TSception in accuracy, MCC, and AUROC. For accuracy, our model improved TSception’s performance in valence score by 0.89\%, arousal score by 1.02\%, and dominance score by 0.51\%, indicating that our model was making more correct predictions overall. For MCC, our model outperforms TSception’s coefficients in valence score by 0.0221, arousal score by 0.0068, and dominance score by 0.0003, suggesting that our model is better at handling both positive and negative classes well, even in imbalanced datasets. Lastly, our model achieves similar improvements across all three scores compared to the TSception model in AUROC, which reflects the ability to discriminate between classes of a model. Our model surpasses TSception in valence score by 0.0084, arousal score by 0.0068, and dominance score by 0.0053.


\section{Discussion}
\label{sec:sec05}

This study presented a novel approach to utilizing DL for EEG-based emotion recognition with visual-auditory stimuli in a consumer-friendly setting, predicting the emotional state of a person in the form of valence, arousal, and dominance scores. Inspired by the TSception model, we used five different ratio coefficients to allow our model to have more diverse representations in the EEG data than previous methods. Moreover, we expanded on the proposed model architecture of TSception, further using a similar approach to the model’s hemisphere kernel to implement a new type of kernel that captures key features from four separate areas of the brain, while still retaining the global and hemisphere kernel from TSception. This approach results in our model consistently outperforming TSception in all valence, arousal, and dominance scores across multiple performance evaluation metrics, including five-fold cross-validation (test accuracy across five folds), precision, recall, F1 score, accuracy, MCC, AUROC, and Cohen’s kappa. Our study also opens up many potential applications in different real-life contexts of EEG-based emotion recognition, as our model was trained on a dataset recorded using a consumer-grade EEG device.

Most EEG-based emotion recognition studies have been using other widely used datasets like DEAP, SEED, and MAHNOB-HCI \cite{_06_Suhaimi2020EEG}, \cite{_32_Roy2019Deep}, \cite{_63_Dadebayev2022EEG}. These datasets, while providing more EEG data with higher quality, use research-grade EEG devices. The use of these devices prevents these studies from developing a system that can be more widely used in real-life contexts, as research-grade EEG devices are extremely costly and take a significant amount of time to set up. The DREAMER dataset recorded EEG data using the Emotiv EPOC system, a consumer-grade EEG system that takes way less time to set up and has more affordability while recording EEG data with satisfactory quality and reproducibility. By training on the DREAMER dataset, our model can better predict emotional states using EEG data from EEG systems with low cost and high user convenience, providing an inexpensive option for hospitals, clinicians, or mental health services with a low budget to install such applications.

With affordability and user convenience in mind, our EEG-based emotion recognition system may be used in a wide range of real-life applications. One of the prime examples of such applications would be monitoring patients’ emotional state during a therapy session. In 2023, one in five people from Gen Z or Millennials (people who were born between 1981 and 2012) are currently treated with psychotherapy as it remains the most popular treatment method for mental disorders \cite{_64_Thriving_Center_of_Psychology_2023}, \cite{_65_Angermeyer2017Public}. As the communication and interaction between the therapist and the clinical patient is the most important aspect of a therapy session \cite{_66_Priebe2008Therapeutic}, therapy experience for individuals with cognitive impairments, especially Alzheimer’s disease, becomes challenging due to them facing difficulties in communication and expressing themselves \cite{_67_Woodward2013Aspects}. Hence, there is a rising need for ways to aid the communication between the therapist and individuals with Alzheimer’s disease in a therapy session. One method that can contribute to this purpose is to develop an automatic emotion recognition system that allows the therapist to understand what state of mind the patient is feeling. By monitoring the valence of a patient, the therapist can better understand their patient’s feelings, detect early warnings of mental problems like depression or anxiety, and then tailor treatments accordingly. During a therapy session that focuses on the use of a stimulus, not only the valence, but the arousal and dominance indication of a patient may provide the therapist with more information regarding the efficacy of the activity and whether a stimulus is working as intended. Music therapy is an evidence-based therapeutic practice that uses music to address the physical, cognitive, and social needs of an individual \cite{_68_Mays2008Treating}. In an active listening activity in a music therapy session, where the patient would listen to a recording of a song, watch a music video, or a live performance of other patients and/or of music therapists. This activity would provide a perfect setting for EEG-based emotion recognition to come into use, as the music therapist would be able to observe whether the activity is eliciting the planned emotion in the patient. Furthermore, the arousal score indicated by the emotion recognition system would be able to suggest the efficacy of the current music therapy activity. Attention is a human’s most basic resource at a psychic level \cite{_69_Csikszentmihalyi2004Materialism}, and thus, many studies exploring the factors affecting the efficacy of music therapy for individuals with dementia place an emphasis on capturing the patient’s attention \cite{_70_Ridder2005Individual}, \cite{_71_barwick2014describing}. The arousal score, which presents the attention level of a person, would be a suitable indicator for the music therapist to adjust their activity in accordance with the score. Another useful application for this system would be in a group music therapy session, where tracking the emotional state of multiple patients at once may become challenging for the music therapist. In particular, the dominance score may be a good indication for the music therapist to recognize if there is a patient who is feeling left out or overwhelmed, which would be suggested by a low dominance score.

Our study is not without its limitations and challenges, and a discussion regarding these limitations would open up future works and ideas. One major challenge that we faced during this study was the lack of data for consumer-friend EEG-based emotion recognition. While the DREAMER dataset was able to provide us with a satisfactory number of samples, the dataset only recorded data from 23 subjects. In the context of a real-life application for an emotion recognition system, it is essential that a model can perform well across many different subjects and different stimuli. For a ML model to achieve that, we hypothesized that it is better for the model to treat the whole recording of a subject with a stimulus as one sample. In the context of the DREAMER dataset, this would mean that the EEG signals recorded while one subject was watching a film clip would count as one sample. However, in the observed EEG datasets during this study, doing this would leave the ML model with too few samples to train on (e.g. the DREAMER dataset would have 414 samples, as it has 23 subjects watching 18 film clips), making convergence extremely unlikely. This is the reason why most EEG-based emotion recognition systems would treat the recorded EEG signals during one time window (this varies in accordance with the recording frequency of the EEG device). Thus, we believe that the field of EEG-based emotion recognition systems with real-life applications would massively benefit from an EEG dataset recorded using a consumer-grade EEG device with emotion evaluations. 

We also encourage future works exploring EEG-based emotion recognition systems to experiment with more unique DL architectures. While the TSception model achieved quality results by designing their architecture to capture the key information from the brain’s activity in the two hemispheres separately, our model managed to achieve consistently better results by separating the EEG data into four areas of the brain. Similar to TSception, our approach was inspired by the brain’s anatomy, specifically the four lobes of the cerebral cortex: the frontal lobe, the parietal lobe, the temporal lobe, and the occipital lobe, even though the electrode placements in the DREAMER dataset did not allow us to design an architecture that would capture this information perfectly. Hence, we highly encourage future works to try novel model architectures, even if they are illogical or if the logic behind them is incomplete. Additionally, future works may explore the performance of our model in a dataset that would better fit for capturing the brain information in the four lobes of the cerebral cortex.

Though this study focuses specifically on the real-life application aspect of EEG-based emotion recognition, we were unable to obtain permission to put the model into real-life use in a suitable setting such as during an active listening activity in a music therapy session. Since modern EEG devices are non-invasive and completely safe with extremely low health risks \cite{_72_2020Cortical}, we suggest testing out such EEG-based emotion recognition systems in a real-life setting in order to better evaluate the shortcomings and practicality of such systems.

As related fields like technology, AI, ML, DL, affective neuroscience, and EEG-based emotion recognition advance in the future, the development and application of these systems should be further explored for their wide range of potential utilization. We call for the creation of more EEG-based emotion recognition datasets that would simulate practical settings for the use of these systems in real life. We hope that this study succeeds in encouraging researchers to experiment with more architectures for such systems and to shift the focus to the development of them in real-life settings, providing easy access to hospitals, clinicians, therapists, and consumers. In the quest for further understanding of the great complexity behind human emotions and using them to improve quality of life, the development of real-life EEG-based emotion recognition systems would ameliorate countless lives.


\section{Conclusion}
\label{sec:sec06}

In this study, we developed a DL model to predict an individual’s emotional state using EEG signals using a multi-scale CNN approach inspired by TSception. By using more ratio coefficients for T kernels and creating a novel type of kernel that would capture the key information of EEG data from four separate areas of the brain, we consistently outperformed the results of the TSception model in all valence, arousal, and dominance scores across a wide range of evaluation metrics, including precision, recall, F1 score, accuracy, MCC, AUROC, and Cohen’s kappa. This result indicates the potential of novel architectures in such systems as well as the need for testing such systems in a real-life setting. We call for further research to experiment with various types of model architectures, as well as the development of an EEG-based emotion recognition system using consumer-grade EEG devices with a considerable number of subjects and stimuli, allowing future development of these systems to better generalize in real-life applications.


\section{Acknowledgement}
\label{sec:sec07}

I would like to thank my advisor, Gia Ngo from the School of Electrical \& Computer Engineering, Cornell University, for his advice and mentorship in the development of this research paper.


\bibliographystyle{ieeetr}
\bibliography{mybibliography}

\end{document}